\documentclass[11pt,a4paper]{article}
\usepackage{times,latexsym}
\usepackage{url}
\usepackage[T1]{fontenc}

\usepackage[acceptedWithA]{tacl2018v2}
\usepackage{amsmath}
\usepackage{multirow}
\usepackage{amssymb}
\usepackage[normalem]{ulem}
\usepackage{graphicx}
\usepackage{tabularx}

\usepackage{float}
\usepackage{algorithm2e}
\usepackage{booktabs} 
\usepackage{tablefootnote} 

\let\Delta\varDelta
\newcommand{\dppl}{$\Delta$\textit{ppl}}

\newcommand{\baseppl}{\textit{$ppl^-$}}
\newcommand{\tgtppl}{\textit{$ppl^+$}}
\newcommand{\basepplx}{\textit{$ppl^-_x$}}
\newcommand{\tgtpplx}{\textit{$ppl^+_x$}}

\newcommand{\Thatnb}{\textit{$\theta^+$}}
\newcommand{\Ttilnb}{\textit{$\theta^-$}}
\newcommand{\Dtilnb}{\textit{$D^-$}}
\newcommand{\Dhatnb}{\textit{$D^+$}}
\newcommand{\DtilnbDel}{\textit{$D^-_{\Delta}$}}
\newcommand{\Dtilnbdel}{\textit{$D^-_{\delta}$}}

\newcommand{\DtilDel}{\textit{$\boldsymbol{\DtilnbDel{}}$}}
\newcommand{\Dtildel}{\textit{$\boldsymbol{\Dtilnbdel{}}$}}
\newcommand{\Ttil}{\textit{$\boldsymbol{\Ttilnb{}}$}}
\newcommand{\That}{\textit{$\boldsymbol{\Thatnb{}}$}}
\newcommand{\Dtil}{\textit{$\boldsymbol{\Dtilnb{}}$}}
\newcommand{\Dhat}{\textit{$\boldsymbol{\Dhatnb{}}$}}

\newcommand{\gleu}{\textrm{GLEU$^+$}}

\newcommand{\ffive}{\textrm{$F_{0.5}$ }}

\usepackage{xspace,mfirstuc,tabulary}

\newif\iftaclinstructions
\taclinstructionsfalse 
\iftaclinstructions
\renewcommand{\confidential}{}
\renewcommand{\anonsubtext}{(No author info supplied here, for consistency with
TACL-submission anonymization requirements)}
\newcommand{\instr}
\fi

\iftaclpubformat 

\else

\fi

\title{Data Weighted Training Strategies for Grammatical Error Correction}

\author{Jared Lichtarge \and Chris Alberti \and Shankar Kumar \\
        Google Research\\
        {\sf \{lichtarge,chrisalberti,shankarkumar\}@google.com}}

\date{}

\begin{document}
\maketitle
\begin{abstract}
Recent progress in the task of Grammatical Error Correction (GEC) has been driven by addressing data sparsity, both through new methods for generating large and noisy pretraining data and through the publication of small and higher-quality finetuning data in the BEA-2019 shared task. Building upon recent work in Neural Machine Translation (NMT), we make use of both kinds of data by deriving example-level scores on our large pretraining data based on a smaller, higher-quality dataset. In this work, we perform an empirical study to discover how to best incorporate delta-log-perplexity, a type of example scoring, into a training schedule for GEC. In doing so, we perform experiments that shed light on the function and applicability of delta-log-perplexity. Models trained on scored data achieve state-of-the-art results on common GEC test sets. 
\end{abstract}

\section{Introduction}\label{sec:introduction}

Grammatical Error Correction (GEC), the task of automatically correcting errors in written text, can be framed as a translation task from `bad grammar' to `good grammar.'
This framing has enabled GEC to borrow models and techniques from the vast literature in machine translation (MT). Neural approaches have dominated recent state-of-the-art advances in GEC, and have been shown to be more effective in direct comparison with non-neural methods~\cite{chollampatt2018multilayer,junczys2018approaching}. 
Nevertheless, GEC continues to pose a challenge for data-reliant neural models given that 
the publicly available training data is relatively limited, with the largest corpus numbering only ~2M examples~\cite{mizumoto-etal-2012-effect}. Therefore, much recent work in GEC has focused on diverse methods to address data sparsity by supplementing available annotated corpora with much larger pretraining data~\cite{ge-etal-2018-fluency, kasewa-etal-2018-wronging,lichtarge-corpora,grundkiewicz-etal-2019-neural,zhao-etal-2019-improving}. A contrasting approach to addressing data sparsity in GEC has been explored in the Building Educational Application (BEA) 2019 Shared Task on Grammatical Error Correction \cite{bea2019}. The task introduced the \emph{Write and Improve} training set, a new high-quality annotated corpus numbering only \textasciitilde{}34k examples (referred to in this work as BEA-19 train), and encouraged exploration of low-resource methods by organizing two tracks specifically for data-restricted competition. Despite the relatively small size, many approaches using the BEA-19 train data achieved better results on common GEC test sets than previous approaches that did not have access to this small but high-quality data~\cite{bea2019}.

In the context of neural MT (NMT), models have been shown to be sensitive to noise in the training data \cite{khayrallah-koehn:2018:WNMT}. While much effort has been dedicated to methods which either filter or downweight noisy pretraining data in NMT~\cite{junczysdowmunt2018dualconditional}, less attention has thus far been paid in GEC. To the best of our knowledge, previously explored techniques for filtering pretraining data in GEC are limited to hand-engineered heuristic cutoffs~\cite{wiked_2014} and n-gram language model filtering~\cite{ge-etal-2018-fluency}. 

Recent work in NMT~\cite{wang-etal-2018-denoising} presents a training technique for scoring the `noise' of training data by employing a much smaller, higher-quality `trusted' dataset. The authors describe a curriculum-style training over data scored by this metric, and demonstrate significant improvements over a baseline. We refer to this score as delta-log-perplexity (\dppl{}).




\section{Contributions of this work}\label{sec:contributions}


This work presents an empirical study of training strategies for GEC in multiple dimensions. Using a standard training setup (without scoring), we explore arrangements of GEC corpora into pretraining and finetuning data, establishing a strong baseline. We then apply data scoring via \dppl{} to the GEC task, demonstrating the value of \dppl{} as a heuristic for example quality. By comparing multiple plausible methods for applying \dppl{}, we gain some insight into the interpretation and practical applicability of the metric. 
We train on the scored data via four simple methods that instantiate different intuitions about how to treat a heuristic score for data quality.
We demonstrate performance gains for various strategies incorporating scoring into the training, and present state-of-the-art results on the CoNLL-14~\cite{ng2014conll} and JFLEG~\cite{napoles2017jfleg} test sets.







\section{Related Work}\label{sec:related}


In recent GEC work, most approaches pretrain on some synthetic data and then finetune on the union of multiple annotated data sources, with some variation in which datasets are included for fine-tuning~\cite{grundkiewicz-etal-2019-neural, lichtarge-corpora}. In a thorough study of incorporating generated pseudo-data into GEC training, \newcite{kiyono-etal-2019-empirical} report that this typical pretrain-finetune setup scales with size of pretraining data better than a setup in which all data is trained on simultaneously. \newcite{choe-etal-2019-neural} describe a `sequential transfer learning' approach in which the pretrained model, finetuned on all available annotated data, is finetuned again separately for each test set.
A thorough review of the GEC field is made by \newcite{wang2020comprehensive}.

Data selection in MT has been performed along two dimensions: domain-relevance and denoising. Multiple researchers~\cite{moore-2010-crossentropylm,axelrod-2011-domaindataselection,vanderwees-2017-dynamicselect} have used the difference in cross-entropy between two language models as a criteria for the selection of in-domain sentences. In contrast, ~\newcite{wang-etal-2018-denoising} and ~\newcite{junczysdowmunt2018dualconditional} have used data selection for denoising. Recently, ~\newcite{wang-etal-2019-cocurriculum}
demonstrate that a co-curriculum training for dynamic selection of data that is both clean and in-domain, can outperform independent selection along each of the two dimensions.

\section{Methods}\label{sec:methods}

\subsection{Delta-Log-Perplexity}
\subsubsection{Background}

\citet{wang-etal-2018-denoising} present a metric defined as the difference in log-probability of an individual training example before and after improving a pretrained model by finetuning on a small trusted dataset.\
\citeauthor{wang-etal-2018-denoising} use this metric to order the pretrain data, and train a new model via a curriculum-style strategy using this ordering. 
In their setup, this metric is interpreted as measuring `noise', describing the change in log probability of an example between a noisy pretrained model and its `denoised' finetuned counterpart. Since log perplexity for an example is the negative of the log-probability, we refer to this score as `delta-log-perplexity'(\dppl)\footnote{Note that \dppl{} is a difference between \emph{log} perplexities, not between the example perplexities themselves.}.

\subsubsection{Calculation}
In the most general case, \dppl ~describes the change in a model's log perplexity for an individual example between two checkpoints in model training. 
If the first checkpoint (with parameterization \Ttil{}) is sampled after model convergence on a \emph{base} dataset \Dtil{}, and the second checkpoint (\That{}), after further finetuning on a second \emph{target} dataset \Dhat{}, then the \dppl{} between those models for a given example (composed of input, output pair $(i,o)$) should suggest which of the datasets the example is more similar to, from the perspective of the successive models \Ttil{} and \That{}.  

\begin{equation}
\Delta\emph{ppl}(i,o;\Ttilnb{},\Thatnb) = 
\log p(o|i;\Ttilnb) - \log p(o|i;\Thatnb)
\end{equation}

In the course of this work, we make use of the \emph{relative ordering} of examples from the scored dataset $\DtilDel$ when sorted by their \dppl{} scores, rather than the actual \dppl{} score values\footnote{This allows us to implement curriculum-style data selection and directly weight examples using the same score.}.
We refer to this quantity as `delta-perplexity-rank':
\begin{small}
\begin{align}
\label{eq:dppl_rank}
\nonumber \Dtilnb_\Delta & = & \{(i,o, \Delta ppl(i,o)) | (i,o) \in \Dtilnb\} \\
\delta ppl(i,o;\Dtilnb_\Delta) & = & 1 -\frac{\%ile\_rank(\Delta ppl(i,o); \Dtilnb_\Delta)}{100}
\end{align}
\end{small}
`\%ile\_rank' refers to percentile rank.
$\delta{ppl}$ has range [0,1], and is computed such that the example with the most negative \dppl{} will have the highest $\delta{ppl}$ score of 1. The median example will have a $\delta{ppl}$ of 0.5. 

\begin{algorithm}
\small
\SetKwInput{KwData}{Input}
\KwData{base dataset $\Dtil$, target dataset $\Dhat$}
\KwResult{\dppl{}-scored base dataset $\DtilDel$}
\quad\quad\quad\quad{}$\delta{ppl}$-scored base dataset $\Dtildel$ \\
$\Ttil{}$ $\gets$ \textit{train new model on $\Dtil$} \\
$\That{}$ $\gets$ \textit{finetune $\Ttil{}$ on $\Dhat{}$} \\
$\DtilDel \gets \{\}$ \\
\For{example $x \in \Dtil{}$}{
    $\basepplx \gets -\log p(x.o|x.i, \Ttil{})$ \\
    $\tgtpplx \gets -\log p(x.o|x.i, \That{})$ \\
    $x.\varDelta{ppl} \gets (\tgtpplx - \basepplx)$ \\
    $\DtilDel \gets \DtilDel \cup{} x$ \\
}
$\Dtildel$ $\gets$ $\{\}$ \\
\For{scored example $x \in \DtilDel$}{
    $x.\delta{ppl} \gets 1 -\frac{\%ile\_rank(x.\varDelta{ppl},\DtilDel)}{100}$ \\
    $\Dtildel \gets{} \Dtildel \cup{} x$ \\
}
\caption{Score base data with $\Delta{ppl}$, and calculate $\delta{\emph{ppl}}$ for each sentence pair. The symbols \emph{x.i} and \emph{x.o} refer to the input and output sequences of the example.}
\label{algo:delta-ppl}
\end{algorithm}

\subsubsection{Explanation}
Any example drawn from \Dhat{} should trivially be expected to have a negative \dppl{} because \That{} has just been trained directly upon the exact example, whereas \Ttil{} has never seen the example before. The negative \dppl{} can be explained by suggesting \That{} has begun to memorize the \textit{specific} examples in \Dhat{}. 

Scoring examples drawn from \Dtil{} reveals the value of the technique; both checkpoints have been trained on \Dtil{} and no example in \Dtil{} was present during further training on \Dhat{}, so the \dppl{} reflects the \textit{general} changes learned during the transition from \Ttil{} to \That{}. Examples from \Dtil{} which are similar to examples from \Dhat{} can be expected to have relatively lower log perplexity for \That{}, and thus lower \dppl{}. Examples from \Dtil{} which are markedly different from those of \Dhat{} should be expected to have higher \dppl{} scores.

While \Dtil{} (\emph{base} data) and \Dhat{} (\emph{target} data) refer to the pretraining and fine-tuning datasets, respectively, in our setup,  we note that these two datasets could be selected according to alternative criteria. The only requirement is that these sets differ in terms of some observable qualitative aspect, for which \dppl{} becomes a heuristic. While in this work we use a target dataset to focus on example quality, it may also be feasible to employ a target dataset that differs from the base data chiefly in domain, and use \dppl{} to negotiate domain transfer.





\subsection{Annealing strategies}
\label{sec:anneal}

When \Dhat{} is selected to be `higher quality' than \Dtil{}, then the \dppl{} scores of examples drawn from \Dtil{} provide a heuristic for example quality. Given a heuristic score for example quality, there are many plausible strategies to incorporate the score into a training schedule. We explore the following schemes:
\textbf{[a]} Filter the pretraining data by discarding examples for which $\delta_{ppl} < k$, where $k$ is a fixed cutoff parameter. \textbf{[b]} Instead of discarding data, down-weight the loss on low-scoring examples during training proportionally to their rank: $\emph{weight}_{x} = \delta{ppl}_{x}$.
A more sophisticated variation of filtering the data is employed by \citet{wang-etal-2018-denoising}: \textbf{[c]} define a curriculum by an exponentially decaying function over training, so that by the end of training, only the best-scoring examples remain in the training data.
\[
include_{x}(\delta{ppl}_{x}, k(t)) =  \begin{cases} 
   1 & \text{if } \delta{ppl}_{x} \geq k(t) \\
   0 & \text{if } \delta{ppl}_{x} < k(t),
  \end{cases}
\]
where $k(t) = 0.5^{\frac{t}{H}}$ for training step $t$ and constant $H$.
To combine the benefits of down-weighting and the curriculum-style annealing, we also implement a mixed strategy \textbf{[d]}:
\[
 weight_{x}(k(t)) = \\
  \begin{cases} 
   1 & \delta{ppl}_{x} \geq k(t) \\
   \delta{ppl}_{x} & \delta{ppl}_{x} < k(t)
  \end{cases} \\ 
\]
where $k(t) = 0.5^{\frac{t}{H}}$ for training step $t$ and constant $H$.

\section{Experiment Setup}\label{sec:setup}
\subsection{Model}
\label{sec:model}
We use the \textit{Transformer} sequence-to-sequence model~\cite{vaswani2017attention}, using the \textit{Tensor2Tensor} open-source implementation with the ``transformer\_clean\_big\_tpu'' setting\footnote{\url{https://github.com/tensorflow/tensor2tensor}}. We use  a 32k word piece dictionary~\cite{schuster2012wordpiece}. For all training stages we use the Adafactor optimizer \cite{shazeer2018adafactor}.

\subsection{Data}

We train on the public version of the Lang-8 corpus~\cite{mizumoto-etal-2012-effect}, the FCE corpus \cite{yannakoudakis-etal-2011-new}, and the Cambridge English Write \& Improve training split described in the BEA-2019 shared task (BEA-19 train) \cite{bea2019}. 

The Lang-8 corpus is scraped from the social language learning website\footnote{\url{https://www.Lang-8.com}}, and is composed of potentially erroneous sentences from English-language-learners with crowd-sourced corrections. The corpus includes many sentence pairs that are noisy or irrelevant to GEC for a variety of reasons. In contrast, FCE\footnote{\url{https://www.cl.cam.ac.uk/research/nl/bea2019st/data/fce_v2.1.bea19.tar.gz}} and BEA-19 train\footnote{\url{https://www.cl.cam.ac.uk/research/nl/bea2019st/data/wi+locness_v2.1.bea19.tar.gz}} are much smaller corpora that have been carefully annotated by a small number of professional annotators. Due to their highly-curated origin, these datasets have a much higher proportion of high-quality GEC-relevant sentence pairs than Lang-8.

For pretraining data, we follow \newcite{lichtarge-corpora} in using a large and noisy corpus of edits crawled from Wikipedia's publicly available revision histories (REV). We also use a similar-sized corpus of sentence pairs, where the target sentences are drawn from Wikipedia, and the source sentences are generated via round-trip-translation through a bridge language (RT) \cite{lichtarge-corpora}. We generate four parallel datasets of equal size by round-trip-translating the same `clean' sequences through four bridge languages\footnote{Japanese, Russian, French, and German, following \cite{lichtarge-corpora}.}.
Both pretraining corpora are further probabilistically corrupted via character-level insertions, deletions, transpositions, and replacements. We corrupt each character of REV, which already contains some `natural' spelling errors, at a rate of 0.003 per character. For the RT data, which does not already have spelling errors, we use a rate of 0.005 per character.

Prior research on GEC has employed the NUCLE corpus~\cite{dahlmeier-etal-2013-building} for model training. Our pilot experiments showed that a model pre-trained on REV/RT yielded similar performance when fine-tuned on either Lang-8 or a combination of Lang-8 and NUCLE. Since both corpora contain corrections of sentences written by non-native speakers, and NUCLE, which has only a fourth as many sentences as Lang-8, did not give additional improvements on top of Lang-8, we decided to exclude the corpus in our experiments to simplify the presentation.

\subsection{Non-Scored Training and Finetuning}
When pretraining, we train the \textit{Transformer} model for 1M steps. We set the learning rate to $0.01$ for the first $10,000$ steps, after which we decrease it proportionally to the inverse square root of the number of steps. When finetuning, we set the learning rate to a constant $3 \times 10^{-5}$. Regardless of the dataset being used, we run finetuning for \textasciitilde{}30 epochs.

\subsection{Scored Training and Finetuning}
When applying the scored training strategies to Lang-8, we discard the base model that was used in calculating the \dppl{} scores (which was trained on: Pretrain $\rightarrow$ Lang-8), and start a new finetuning run on the scored Lang-8, from a model initialized on the same pretraining data.

When applying our scored training strategies to the much larger pretraining data, rather than
start the model from random initialization and repeat 1M steps of training, we \emph{continue} training from the 1M checkpoint of the base model and train on the scored data for an additional 100,000 steps (using the same pretraining settings). 

\subsection{Evaluation}
In the course of our experiments, we evaluate on the development set of the BEA-2019 shared task (BEA-19 dev), which includes examples from both W\&I and the LOCNESS corpus \cite{granger1998}, using the ERRANT scorer \cite{bryant-etal-2017-automatic}. In our analysis (Section \ref{sec:analysis}), we report on BEA-19 test, with scores provided by the official Codalab of the BEA-2019 task\footnote{\url{https://competitions.codalab.org/competitions/20229}}. We also report on the popular GEC evaluation corpora: CoNLL-2014~\cite{ng2014conll} and JFLEG~\cite{napoles2017jfleg,heilman2014}, for which we report \ffive with the $M^2$ scorer~\cite{dahlmeier2012better} and the \gleu metric~\cite{napoles2016gleu} respectively. For BEA-19 dev and BEA-19 test, following the conventions of the shared task, we post-processed the model output with the spaCy tokenizer\footnote{\url{https://spacy.io/}}. 

For decoding, we use iterative decoding \cite{lichtarge-corpora} with a beam size of 4.
For each reported test result, we select the model checkpoint, set the number of decoding iterations, and tune a scalar identity threshold based on performance on the corresponding development sets. Ensemble decoding is computed using the average~\cite{cromieres16} of the logits of multiple identical \textit{Transformers}, trained separately.

\section{Experiments}\label{sec:experiments}
\subsection{Standard training}

\begin{table}[h]
  \footnotesize
  \centering
  \begin{tabular}{c|c|c}
    \toprule
    Corpus & sentences & words \\
    \midrule
    FCE & 28K & 432K \\ 
    BEA-19 train & 34K & 560K \\
    Lang-8 & 1.9M & 25.0M \\
    Wiki revisions (REV) & 170M  & 4.1B  \\
    Wiki roundtrip-translated (RT)\footnotemark{} & 170M  & 4.1B  \\
    \bottomrule
  \end{tabular}
  \caption{Training datasets. \emph{Wiki} refers to Wikipedia.  }
  \label{tab:data}
\end{table}
\footnotetext{For each of the four bridge languages. The `clean' target sentences are the shared between the four.}

The datasets presented in Table \ref{tab:data} can be sorted into three categories by their relative quality. REV and RT are noisiest, with most data not appearing relevant to GEC. FCE and BEA-19 train are cleanest, as they are professionally annotated. Lang-8 occupies a middle ground, as the data, which is largely relevant to GEC but scraped from a crowd-sourced medium, does not rise to the standard of professional annotation. In light of this, we combine the single REV dataset with each of the four RT datasets to produce four large pretraining datasets, each containing half Wiki revisions and half round-trip translated data (PRE). All experiments are run for each of these merged datasets, and all reported figures are the average of those four models. We also merge the FCE and BEA-19 train into a single finetuning set, which we refer to as `BEA-FCE' (BF). 

We explore three training schemes: including Lang-8 with the higher-quality annotated data, including Lang-8 with the pretraining data, and a two-stage finetuning scheme, with Lang-8 as the intermediate step.
\begin{table}[!htbp]
  \centering
  \footnotesize
  \begin{tabular}{ccc}
  \toprule
    \multicolumn{2}{c}{Training Data} & BEA-19 dev \ffive \\
    \midrule
    \multirow{2}{*}{1} & PRE & 24.0 \\
      & $\rightarrow$  (Lang-8 $\cup$ BF) &	46.3 \\
     \midrule
     \multirow{2}{*}{2} & (PRE $\cup$ Lang-8) &	32.4 \\
     & $\rightarrow$ BF & 51.4 \\
     \midrule
     \multirow{2}{*}{3} & PRE $\rightarrow$ Lang-8 & 42.5 \\
     & $\rightarrow$ BF &	51.5 \\
     \bottomrule
  \end{tabular}

  \caption{Comparing pretrain-finetune arrangements. The arrow indicates finetuning.}
  \label{tab:standard-training}
\end{table}  

\subsection{Applying delta-log-perplexity}

For experiments [2] and [3] of the standard training setup (Table \ref{tab:standard-training}), we apply delta-log-perplexity scoring. For the multi-stage finetuning setup, we explore arrangements of base ($\boldsymbol{\Dtil}$) and target ($\boldsymbol{\Dhat}$) datasets that ensure that $\boldsymbol{\Dhat}$ is smaller and higher-quality than $\boldsymbol{\Dtil}$. For these experiments, we use the soft-weighting training strategy ([b] in Sec~\ref{sec:anneal}), as it is has no tunable hyper-parameters and does not discard any data. Results are shown in Table \ref{tab:soft-weighted}.

\begin{table}[!htbp]
  \centering
  \footnotesize
  \begin{tabular}{cccc}
   \toprule
    & \multirow{2}{*}{Training Data} &  \multicolumn{2}{c}{BEA-19 dev} \\ \cline{3-4}
    & & \ffive & $\Delta$ vs unscored \\
    \midrule
    \multirow{2}{*}{A} & $(\textbf{PRE $\cup$ Lang-8})_{BF}$  & 44.9 & +12.5 \\ 
      & $\rightarrow$  BF  & 51.8 & +0.4 \\ 
    \midrule
    \multirow{3}{*}{B} & $\textbf{PRE}_{BF}$  & 37.0 & +6.8 \\ 
     & $\rightarrow$ Lang-8 & 43.3 & +0.8 \\ 
     & $\rightarrow$ BF & 51.7 & +0.2 \\ 
    \midrule
    \multirow{3}{*}{C} & PRE & 24.0 & --- \\ 
     & $\rightarrow$ $\textbf{Lang-8}_{BF}$ & 47.2 & +4.7 \\ 
     & $\rightarrow$ BF & 51.9 & +0.4 \\
    \midrule
    \multirow{2}{*}{D} & $\textbf{PRE}_{BF}$ $\rightarrow$ $\textbf{Lang-8}_{BF}$ & 48.0 & +5.5 \\
     & $\rightarrow$ BF & 52.3 & +0.8 \\
    \bottomrule

  \end{tabular}
  \caption{Comparing scoring arrangements. Bold indicates a base dataset for which \dppl{} scores have been calculated, the subscript denotes the target dataset used. e.g. In \emph{A}, the scores are calculated for  $\textbf{PRE $\cup$ Lang-8}$ using BF as the target. All scored datasets are trained via \emph{soft} down-weighting. The final column indicates the change in \ffive over the unscored setup at the same training stage (absolute values in Table \ref{tab:standard-training}).}
  \label{tab:soft-weighted}
\end{table}

\begin{table*}[!htbp]
  \centering
  \footnotesize
  \begin{tabular}{cc|ccccc}
   \toprule
    & \multirow{2}{*}{Training Data} & \multicolumn{5}{c}{Training Strategy} \\ 
    & & \emph{unscored} & \emph{   hard   } & \emph{    soft    } & \emph{hard-cclm} & \emph{soft-cclm} \\
    \midrule
    \multirow{2}{*}{i} 
    & $\textbf{PRE}_{BF}$ (\emph{soft}) $\rightarrow$ $\textbf{Lang-8}_{BF}$* 
    & 43.3 & 49.0 & 48.0 & 45.8 & 47.9 \\ 
    & $\rightarrow$ BF &
    51.7 & 52.1 & 52.3 & 51.8 & 52.4 \\
    \midrule
    \multirow{3}{*}{ii} 
    & $\textbf{PRE}_{BF}$*
    & 24.0 & 45.7 & 37.0 & 47.7 & 36.9 \\ 
    & $\rightarrow$ $\textbf{Lang-8}_{BF}$ (\emph{soft})
    & 42.5 & 48.1 & 48.4 & 48.6 & 48.0 \\ 
    & $\rightarrow$ BF &
    51.5 & 51.8 & 52.4 & 52.3 & 52.2 \\
	\bottomrule

  \end{tabular}
  \caption{Comparing training strategies for $\textbf{PRE}_{BF}$, and $\textbf{Lang-8}_{BF}$, following setup (D) in Table \ref{tab:soft-weighted}. The asterisk indicates the training stage that is being varied in each experiment. In (ii) all models are finetuned on $\textbf{Lang-8}_{BF}$ using the \emph{soft} strategy. The \emph{hard} strategies filter out all examples with positive \dppl{}, which leaves 37\% of the dataset remaining for both $\textbf{PRE}_{BF}$, and $\textbf{Lang-8}_{BF}$. The curriculum strategies anneal down to the best 5\% of the dataset, following \cite{wang-etal-2018-denoising}.}
  \label{tab:strategies}
\end{table*}

\subsection{Training with scored examples}

Given a set of training data for which each example has an associated heuristic `quality' score, there are many plausible options for incorporating that score into a training schedule. For the best-performing scoring arrangement, [D] in Table \ref{tab:soft-weighted}, we repeat the scored training stage in order to compare the following strategies for incorporating scores in training.
\begin{table}[!htbp]
  \footnotesize
  \centering
  \begin{tabular}{ccc}
    \textbf{(a)} & \emph{hard} & Filter by preset rank-score cutoff \\
    \textbf{(b)} & \emph{soft} & Down-weight loss by rank-score \\
    \textbf{(c)} & \emph{hard-cclm} & Curriculum-style filtering \\
    \textbf{(d)} & \emph{soft-cclm} & Curriculum-style down-weighting \\
  \end{tabular}
\end{table}

\noindent Results are shown in Table~\ref{tab:strategies}. We note that ~\newcite{wang-etal-2018-denoising} employed the \emph{hard-cclm} strategy for noise-filtering in NMT.

\section{Analysis}\label{sec:analysis}
\subsection{Understanding \dppl{} scores}
\begin{figure*}[!htbp]
	\centering
    \includegraphics[height=2.3in]{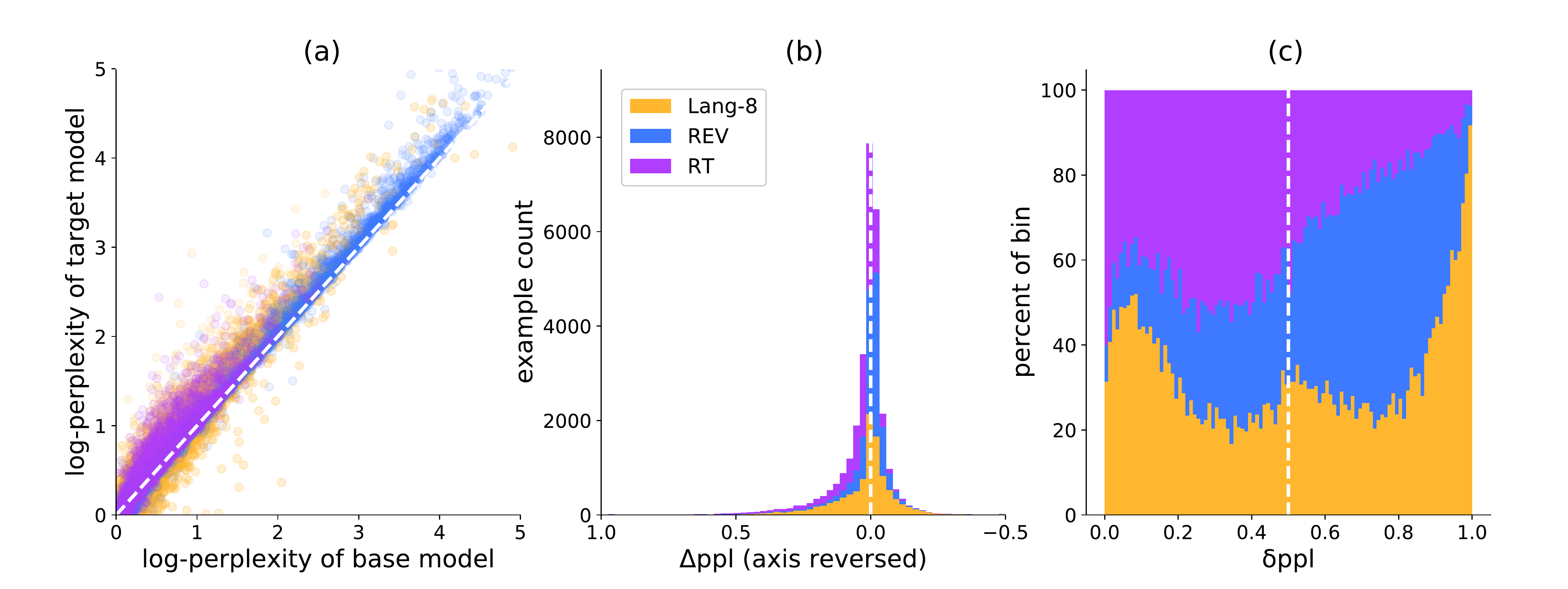}
        \caption{A comparison of the log-perplexity of base and target models (a), the corresponding histogram across the \dppl{} axis (b), and the relative proportions of the three datasets in each $\delta{ppl}$ percentile (c), for 30k examples sampled from $\textbf{PRE $\cup$ Lang-8}_{BF}$ such that 10k examples were selected from REV, RT and Lang-8 respectively. The histogram (b) x-axis has been reversed to align the `best' examples (with the lowest \dppl{}) towards the right, copying the alignment of the $\delta{ppl}$ plot (c); for the scatter plot (a), the best examples are towards the bottom-right. The $\delta{ppl}$ scores shown (c) are the values actually employed by the various training strategies.}
	\label{fig:three_images}
\end{figure*}
\begin{table*}[!htbp]
  \centering
  \footnotesize
  \begin{tabular}{ccccc}
    Dataset & & Example & $\tgtppl$ & \dppl{} \\
    \toprule
    \multirow{8}{*}{REV} 
    & a & It \emph{\sout{comprises} \textbf{gives birth to}} 3 genera. & 2.44 & -0.25 \\
    & b & They \emph{\sout{also}} can \emph{\textbf{also}} live in desert and forest areas.  & 0.1 & -0.14 \\ 
    & c & It included 10 tracks, half \emph{\textbf{of}} them with Joe on vocals. & 0.07 & -0.05 \\\cline{2-3}
     & \multirow{2}{*}{d} & The \emph{\textbf{threee churches in the latter}} parish \emph{\textbf{, at Rathgaroguie, Cushintown}} & \multirow{2}{*}{3.89}& \multirow{2}{*}{0.06}\\ 
     & &  \emph{\textbf{and Terrerath, cater for} \sout{has}} a population of approximately 2500.  &  & \\ \cline{2-3}

     & \multirow{2}{*}{e} & Browsing by subject, for example, \emph{\sout{is} \textbf{was}} possible as \emph{\sout{is} \textbf{was}} restricting searches  & \multirow{2}{*}{0.42}& \multirow{2}{*}{0.12}\\ 
     & & to a particular subject or type of resource.   &  & \\  \cline{2-3}
     & f & She drove a blue \emph{\textbf{Ford}} SUV.  & 1.36 & 0.14 \\ 
     & g & \emph{\sout{The circle is complete.}} Fr.  & 2.65 & 2.06 \\ 

     \midrule
    \multirow{7}{*}{RT} 
    & h &  In winter, \emph{\textbf{the}} sport was hockey. & 0.1 & -0.2 \\
    & i & Nearly a thousand people \emph{\sout{was} \textbf{were}} injured. & 0.01 & -0.16 \\
    & j & This section \emph{\textbf{provides}} only \emph{\sout{provides}} a brief overview of some translated versions. & 0.16 & -0.08 \\
    & k & The sets are now \emph{\sout{depleted} \textbf{out of print}}. & 0.12 & 0.06 \\
    & l &  In 1902 \emph{\sout{,}} they held a garden party on the grounds of the Rose Bay Cottage. & 0.23 & 0.1 \\
    & m & This \emph{\textbf{meant a reduction of the runtime by}} \emph{\sout{resulted in a}} 25 minute\emph{\textbf{s}} \emph{\sout{run time reduction}}.  & 1.19 & 0.15 \\
    & n &  \emph{\sout{The bad case was} \textbf{Adverse weather is}} the third \emph{\textbf{largest}} cause of accidents.   & 2.0 & 0.5 \\
    
    \midrule
    
    \multirow{7}{*}{Lang-8} 
    & o &  Please check \emph{\sout{it}} whether \emph{\textbf{the}} way of speaking is right. & 0.09 & -0.18 \\
    & p & So, can't government make \emph{\textbf{up}} for holiday gaps & 0.43 & -0.12 \\
    & q &  I really enjoyed watching the movie \emph{\textbf{,}} although I never read the manga. & 0.14 & -0.08 \\
    & r &  I am \emph{\sout{worry}} \emph{\textbf{worried}} about their \emph{\sout{damages of mind} \textbf{mental well-being}}. & 1.03 & -0.003 \\
    & s &  I always wake up 6 \emph{\sout{AM every days} \textbf{a.m.everyday}} and then I go to college. & 1.05 & 0.11 \\
    & t &  \emph{\sout{First} \textbf{The first}} time, He \emph{\sout{applogized} \textbf{apologized}} to me, & 0.5 & 0.12 \\
    & u &  I often use \emph{\sout{the}} google \emph{\sout{translation} \textbf{translator}}. & 1.33 & 0.27 \\

    \bottomrule
  \end{tabular}
  \caption{Examples from $\textbf{PRE $\cup$ Lang-8}_{BF}$. Italicized text represents differences between source and target. Strikethroughs represent deletions and bold text represents insertions.}
  \label{tab:examples}
\end{table*}

Training a model on $\textbf{PRE $\cup$ Lang-8}_{BF}$ ([A] in Table \ref{tab:soft-weighted}) achieves a +12.5 \ffive gain over a model trained on the same unscored dataset, and outperforms a model trained on PRE $\rightarrow$ Lang-8 by +2.4 \ffive on BEA-19 dev ([3] in Table \ref{tab:standard-training}). Figure \ref{fig:three_images} explores the characteristics of the \dppl{} scores for the merged dataset, with examples labeled by their original source dataset (REV, RT, or Lang-8).

The scatter-plot (a) offers some insight into how \dppl{} works. Strikingly, all data clusters tightly around the diagonal on which \dppl{}=0. Almost all examples with negative \dppl{} also have low $\tgtppl$ as well. Variance in \dppl{} between examples is much less than variance in $\tgtppl$.
The scatter-plot yields distinct shapes for each of the datasets, and the percentile-rank plot (c) (which depicts the relative proportions of each dataset per percentile bin) shows that the datasets have drastically different scoring profiles. Lang-8, RT and REV have 52\%, 30\%, and 66\% examples with negative (good) \dppl{} respectively, and Lang-8 carries a disproportionate share of the most extreme examples in either direction. Inspecting individual examples helps to elucidate why.

In Table \ref{tab:examples} we draw individual examples from $\textbf{PRE $\cup$ Lang-8}_{BF}$ alongside their $\tgtppl$ and \dppl{} scores. The examples exhibit some characteristics particular to the methodology of their origination. 

\subsubsection{Wikipedia Revisions}
Some of the REV examples [d,f,g] demonstrate the shortcomings of the dataset; significant additions or deletions of information with no grammatical content. While most such examples have positive (bad) \dppl{}, it is noteworthy that example [d], which seems catastrophically out-of-domain, has a better \dppl{} than [e], which simply changes the tense of the sentence. $\tgtppl$ is much higher for examples that have significant information change. This explains why the REV data in the scatter-plot extends thinly along the \dppl{}=0 diagonal; REV contains many examples with information change, for which both source and target are grammatically correct. For these examples, absolute value of both $\tgtppl$ and $\baseppl$ is large, but the change in \dppl{} is relatively small. This demonstrates a shortcoming of using only \dppl{} as a heuristic for example quality: REV has a higher percentage of `good' examples than Lang-8 according to \dppl{}, but many of those examples actually have large $\tgtppl$, and do not capture grammatical changes. Example [a] illustrates a related failure case; it has high $\baseppl$, but according to \dppl{} alone, is the `best' example in the table.

\subsubsection{Roundtrip-translations}
The roundtrip-translated data does not suffer from large information changes, except when the meaning is so garbled as to produce a semantically irreconcilable sequence, as in [n]. As a result, the distribution of RT examples has lower $\tgtppl$ than that of REV. However, many examples include re-arrangements or re-phrasings that are out of scope for the task of GEC [k, m]; of the 10k sampled examples, only 30\% have `good' (negative) \dppl{}. Interestingly, in example [l], passing a sequence through two translation models introduced a reasonably placed comma in what should have been the `corrupted' source sequence; removing this comma yields a bad \dppl{} score.

\subsubsection{Lang-8}
Most Lang-8 examples, for better or worse, do involve grammatically relevant changes to the source sequence. Lang-8 contains many sentence pairs that contain some bad or awkward changes, and these examples perform poorly according to \dppl{} [s, u]. Interestingly, partial corrections, even apparently good ones, also perform poorly [t]. This may be a result of the relatively complete nature of the corrections made in BF, in which few if any target sequences appear to need further correction.

\subsection{Training strategies}


The scored training strategies (Table~\ref{tab:strategies}) explore approaches to making use of an example-level quality heuristic that accommodate distinct intuitions about how to treat the data. Filtering out examples beforehand (\emph{hard}) follows the intuition that bad examples only hurt performance and should be excluded. Down-weighting the loss (\emph{soft}) modifies the relative importance of examples, but avoids throwing any out, maintaining the value of having a large dataset. The `curriculum'-style counterparts of each apply the same logic, while incorporating (albeit in a hard-coded manner) the intuition that the value of some examples changes over the course of training. 

It is worthwhile to note that the optimal strategy, even amongst these simple hard-coded strategies, is a function of the characteristics of the dataset in question. The \emph{hard-cclm} strategy is worst for $\textbf{Lang-8}_{BF}$, where it gradually isolates a small portion of an already small dataset, but is best for $\textbf{PRE}_{BF}$, which is so large that 5\% of the dataset is still considerable. Also, much of what is lost in the `bad' portion of $\textbf{PRE}_{BF}$ is lower-quality data than that which exists in $\textbf{Lang-8}_{BF}$, which may explain both why \emph{hard-cclm} does so well for $\textbf{PRE}_{BF}$ and why \emph{soft-cclm}, which does not throw out the large portion of bad examples, does relatively poorly.

The \emph{hard} strategy outperforms both \emph{soft} and \emph{soft-cclm} for the first stage of both experiments, but the advantage disappears following finetuning on BF. This suggests that cutting out the `worst' examples entirely, while beneficial in the scored training stage, may prevent a sort of regularization that is beneficial to the ultimate finetuned model.

That all strategies similarly out-perform the baseline suggests that \dppl{} is a robust heuristic for quality; that all are simple and un-tuned to the data suggests that there remains headroom for more sophisticated training strategies to do even better.

\subsection{Scoring with less target data}
We observe that scoring any combination of lower-quality datasets using BF as the target data leads  to large improvements over unscored pretraining models, and modest performance gains over those unscored models after finetuning (Table \ref{tab:soft-weighted}).  
\begin{table}[!htbp]
  \footnotesize
  \centering
  \begin{tabular}{c|c|c}
    \toprule
    Dataset proportion & examples & learning rate \\
    \midrule
    full & 60011 & $3 \times 10^{-5}$ \\ 
    \textasciitilde{}1/2 & 29998 & $3 \times 10^{-5}$ \\ 
    \textasciitilde{}1/4 & 15121 & $25 \times 10^{-6}$ \\ 
    \textasciitilde{}1/8 & 7608 & $1 \times 10^{-7}$ \\ 
    \textasciitilde{}1/16 & 3749 & $1 \times 10^{-7}$ \\ 
    \textasciitilde{}1/32 & 1841 & $1 \times 10^{-7}$ \\ 
    \textasciitilde{}1/64 & 905 & $1 \times 10^{-7}$ \\ 
    \bottomrule
  \end{tabular}
  \caption{Successive halves of the BF dataset used in Figure \ref{fig:size-of-target}. Proportion of FCE and BEA-19 train is held constant during down-sampling. Learning rates are tuned based on the test set of the CoNLL-2013 shared task.}
  \label{tab:learning-rates}
\end{table}

\begin{figure*}[!htbp]
	\centering
	\includegraphics[height=2.3in]{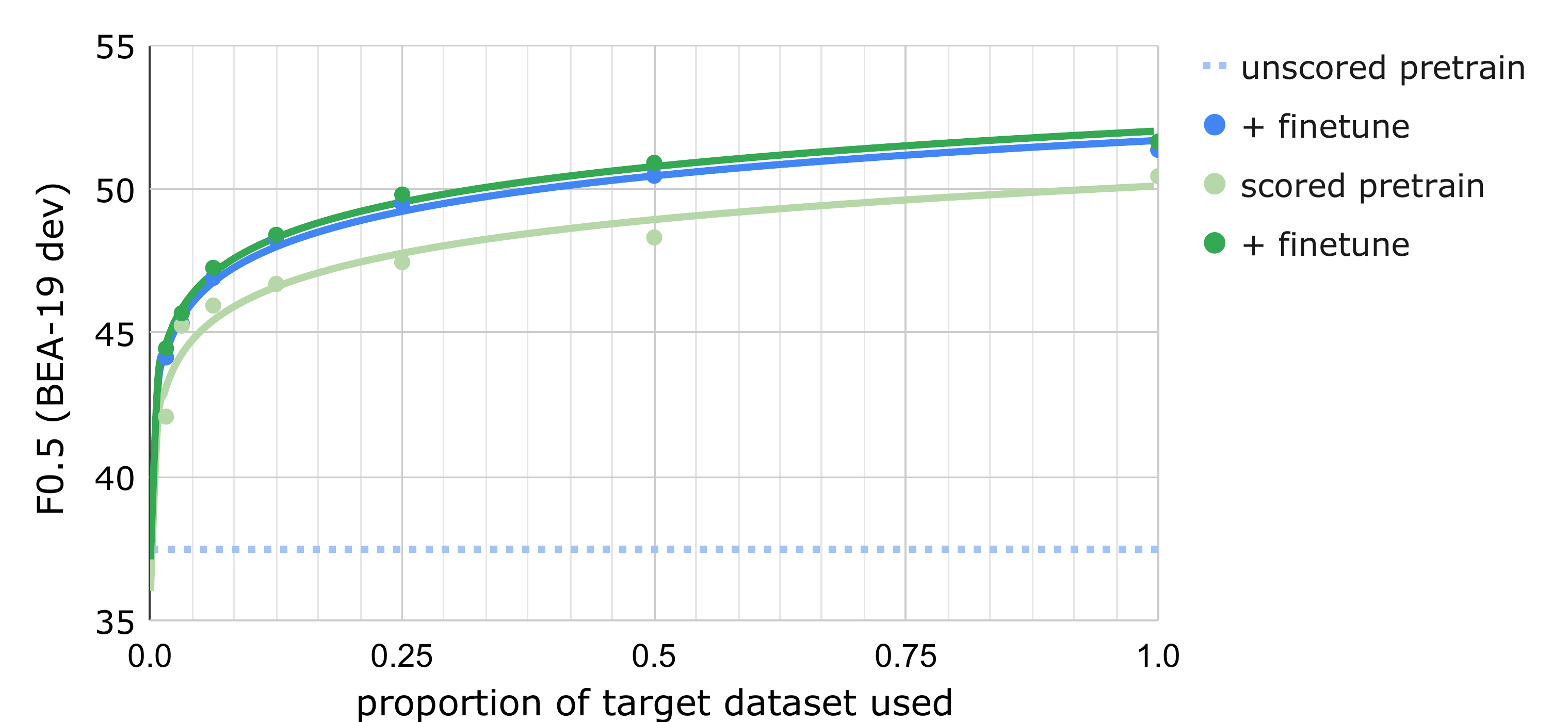}
        \caption{Performance of scored and unscored  pretraining and finetuning as a function of proportion of the target dataset used. The pretraining dataset is (PRE $\cup$ Lang-8). Full BF dataset is shown at the far right (n=60011). Each smaller dataset is a randomly halved subset of the last, with proportion of BEA-19 train / FCE examples held constant. The smallest subset, (BF randomly halved six times) has 905 examples. Logarithmic lines of best fit are shown.}
	\label{fig:size-of-target}
\end{figure*}

We now explore how each of these effects varies as a function of the target data size. For the scoring setup with the largest relative gains over unscored pretraining ([A] in Table~\ref{tab:soft-weighted}), we repeat the same experiment multiple times, but using nested subsets of BF for both scoring and finetuning, each half the size of the previous one. While halving the datasets, we maintain the ratio of BEA-19 train and FCE data within each subset. Because using the same finetuning learning rate would quickly overfit for the smaller datasets, learning rates were tuned for each subset using the test set of the CoNLL-2013 shared task~\cite{ng-conll-2013} (Table~\ref{tab:learning-rates}).

All models are trained via the \emph{hard-cclm} strategy, which, prior to finetuning, significantly outperforms other strategies for training on scored pretraining data (section `ii' in Table \ref{tab:strategies}). Results are shown in Figure \ref{fig:size-of-target}.

\subsection{Understanding the benefits of scoring}

The marginal benefit of scoring the pretraining data yields a drastic performance gain over unscored pretraining, even for very small amounts of target data (see Figure~\ref{fig:size-of-target} and Table \ref{tab:soft-weighted}). This pretrain gain reflects the value of obliquely incorporating the information of the target dataset into the pretraining data via \dppl{} scoring. Because finetuning on the target dataset directly incorporates that same information again, this gain is diminished once the scored models are finetuned (see "$\Delta$ vs unscored" column in Table \ref{tab:soft-weighted}). However, the benefits of finetuning are limited by over-fitting to the finetuning dataset, which is likely to occur given that it is substantially smaller ($\approx$ 1M words) than pretraining data ($\approx$ 8B words). Thus the scored pretrained model, which has already incorporated some of the information of the target dataset without yet having seen any of the specific examples therein, is able to make better use of the finetuning set before the harm of over-fitting outweighs the benefit of further training. This difference explains why even after finetuning, the models with scored training stages outperform the unscored models, though by less than if directly comparing the scored and unscored stages themselves.

In Figure \ref{fig:size-of-target}, the marginal benefit of scoring for the 30k dataset size is +0.5 $F_{0.5}\text{,}$ compared to +0.9 \ffive for doubling the size of the finetuning data (without scoring). For tasks constrained by the availability of high-quality data, and for which labeling costs are high, scoring noisy pretraining data may be a thrifty path to performance gains.

\subsection{Test set results}
\begin{table*}[!htbp]
  \centering
  \footnotesize
  \begin{tabular}{cc|ccc|ccc|c}
  \toprule
    &\multirow{2}{*}{Training Strategy} & \multicolumn{3}{c}{BEA-19 test} & \multicolumn{3}{c}{CoNLL-14 test} & JFLEG test  \\ 
    & & Prec. & Rec. & \ffive (ERRANT) & Prec. & Rec. & \ffive ($M^2$) & \gleu  \\ 
    \midrule
	\multirow{4}{*}{\textbf{unscored}}
	& PRE 
	& 35.7 & 41.7 & 36.8 &	44.6	&	36.2	&	42.6	&	54.1 \\
    & $\rightarrow$ Lang-8 
    & 62.7 & 52.4 & 60.3	&	64.0	&	42.8	&	58.3	&	62.5 \\
    & $\rightarrow$ BF 
    & 67.4 & 61.7 & 66.1	&	67.6	&	44.3	&	61.1	&	63.6 \\
	& \emph{ensemble}
    & 74.1 & 64.3 & 71.9	&	72.6	&	46.7	&	65.3	&	64.7 \\
    \midrule
	\multirow{7}{*}{\textbf{scored}}
    & $\textbf{PRE}_{BF}$ (\emph{soft}) 
    & 56.6 & 47.1 & 54.4	&	61.6	&	38.2	&	54.8	&	59.4 \\
    & $\rightarrow$ $\textbf{Lang-8}_{BF}$ (\emph{soft}) 
    & 68.0 & 57.8 & 65.7	&	68.6	&	44.7	&	62.0	&	63.7 \\
    & $\rightarrow$ BF
    & 67.6 & 62.5 & \textbf{66.5}	&	69.4	&	43.9	&	\textbf{62.1}	&	\textbf{63.8} \\
	& \emph{ensemble}
    & 75.4 & 64.7 & \textbf{73.0}	&	74.7	&	46.9	&	\textbf{66.8}	&	64.5 \\ \cline{2-9}
    & $\textbf{PRE}_{BF}$ (\emph{soft}) $\rightarrow$ Lang-8
    & 64.1 & 52.2 & 61.3	&	66.0	&	41.8	&	59.2	&	62.5 \\
    & $\rightarrow$ BF
    & 66.8 & 61.5 & 65.7	&	68.3	&	45.4	&	62.0	&	63.6 \\
	& \emph{ensemble}
    & 71.7 & 67.4 & 70.8	&	71.2	&	49.9	&	65.6	&	\textbf{64.9} \\
	\bottomrule
  \end{tabular}
  \caption{Test set evaluation results. For each test set, the finetuning checkpoint selected, the identity-correction threshold, and the number of rounds of iterative decoding are tuned to the respective dev sets. BEA-19 test results are provided via the Codalab competition website of the BEA-2019 shared task. Each non-\emph{ensemble} row represents the average of four models, whose construction is described in Section~\ref{sec:experiments}. The \emph{ensembles} combine the four models from the preceding row.}
  \label{tab:testsets}
\end{table*}
\begin{table*}[!htbp]
  \centering
  \footnotesize
  \begin{tabular}{cc|ccc}
  \toprule
    & & BEA-19 test & CoNLL-14 test & JFLEG test  \\ 
    & & \ffive (ERRANT) & \ffive ($M^2$) & \gleu  \\ 
    \midrule
	\multirow{5}{*}{\textbf{single model}}
    & \cite{kiyono-etal-2019-empirical} & 64.2 & 61.3 & 59.7 \\
    & \cite{lichtarge-corpora} & --- & 56.8 & 61.6 \\
    & \cite{xu-etal-2019-erroneous} & --- & 60.9 & 60.8 \\
    & \cite{omelianchuk2020gector} & 72.4 & 65.3 & --- \\ 
    \cline{2-5}
	& \emph{this work - unscored} & 66.1 & 61.1 & 63.6 \\
	& \emph{this work - scored} & 66.5 & 62.1 & 63.8 \\
    \midrule
	\multirow{8}{*}{\textbf{ensemble}}
    & \cite{choe-etal-2019-neural} & 69.1 & 60.3 & --- \\
    & \cite{DBLP:journals/corr/abs-1807-01270}  & --- & 61.3 & 62.4 \\
	& \cite{grundkiewicz-etal-2019-neural} & 69.5 & 64.2 & 61.2 \\
    & \cite{kiyono-etal-2019-empirical} & 70.2 & 65.0 & 61.4 \\
    & \cite{lichtarge-corpora} & --- & 60.4 & 63.3 \\
    & \cite{xu-etal-2019-erroneous} & 66.6 & 63.2 & 62.6 \\
    & \cite{omelianchuk2020gector} & \textbf{73.7} & 66.5 & --- \\ 
    \cline{2-5}
	& \emph{this work - unscored} & 71.9 & 65.3 & 64.7 \\
	& \emph{this work - scored} & 73.0 & \textbf{66.8} & \textbf{64.9} \\
	\bottomrule
  \end{tabular}
  \caption{Comparison of test set evaluation results to prior work, showing the best reported result for each test set in each cited work. Cited values for different test sets do not necessarily represent the same model.}
  \label{tab:comparison}
\end{table*}
We evaluate our best unscored and scored systems at all stages of training on BEA-19 test, CoNLL-14 and JFLEG. Results are shown in Table \ref{tab:testsets}. Results for BEA-19 test are provided by the official Codalab competition of the BEA-2019 shared task, where this work qualifies as \textit{Unrestricted} due to its reliance on additional parallel data like the Wikipedia revisions pretraining dataset. Because the most competitive results in the BEA-2019 task were submitted to the \textit{Restricted} track, the results of this work are not perfectly comparable to most recent and competitive GEC publications. 
Additionally, many of the cited works make use of the NUCLE dataset \cite{dahlmeier-etal-2013-building}, which was not used in this work. Nonetheless it is useful to contextualize the results within the scope of recent progress in GEC. A comparison to recent prior work is made in Table \ref{tab:comparison}. This work achieves state-of-the-art results for the JFLEG and CoNLL-14 test sets, and obtains competitive results on BEA-19 test. 


\section{Future Work} 

The huge jump in performance between unscored and scored pretraining data demonstrates the possibility of making much more effective use of large and noisy datasets through the incorporation of example-level quality scores. While \dppl{} is one such score, there is significant room for improvement, as seen in the example-level analysis in Section \ref{sec:analysis}. Other methods for scoring individual examples should be explored.

In our scored training, we have presented hard-coded training strategies selected for their simplicity. These un-tuned strategies are easy to implement, but do not represent optimal uses of an example-level heuristic score. The fact that there is such variability between them in the two experiments of Table \ref{tab:strategies} suggests that training methods that are sensitive to the particularities of the scored dataset and the model may be able to make much better use of the same scored data. For example, a training scheme that, during training, dynamically decided which data to include or exclude (or how to weight the included data) could be expected to outperform our hard-coded strategies and hyperparameters. A training strategy along these lines has been implemented successfully by \newcite{kumar2019reinforcement} for NMT.

These two complementary directions of future work, the development of new example-level quality heuristics, and the techniques to apply them in scored training, present an intriguing path for future exploration.






\section*{Acknowledgements} 
The authors would like to thank Felix Stahlberg, and the three anonymous reviewers, for their helpful comments.

\bibliography{tacl2018}
\bibliographystyle{acl_natbib}

\end{document}